\documentclass[a4paper,conference]{IEEEtran}
\ifCLASSINFOpdf
\else
\fi
\usepackage{bm}
\usepackage{amssymb}
\usepackage[normalem]{ulem}
\useunder{\uline}{\ul}{}
\begin{document}
%
\title{Self-Attention Neural Bag-of-Features}

\author{\IEEEauthorblockN{Kateryna Chumachenko$^1$, Alexandros Iosifidis$^2$ and Moncef Gabbouj$^1$}
\IEEEauthorblockA{$^1$Department of Computing Sciences, Tampere University, Tampere, Finland\\
$^2$Department of Electrical and Computer Engineering, Aarhus University, Aarhus, Denmark\\
Emails: \{kateryna.chumachenko,moncef.gabbouj\}@tuni.fi, ai@ece.au.dk}
}


%


\maketitle

\begin{abstract}
In this work, we propose several attention formulations for multivariate sequence data. We build on top of the recently introduced  2D-Attention and reformulate the attention learning methodology by quantifying the relevance of feature/temporal dimensions through latent spaces based on self-attention rather than learning them directly. In addition, we propose a joint feature-temporal attention mechanism that learns a joint 2D attention mask highlighting relevant information without treating feature and temporal representations independently. The proposed approaches can be used in various architectures and we specifically evaluate their application together with Neural Bag of Features feature extraction module. Experiments on several sequence data analysis tasks show the improved performance yielded by our approach compared to standard methods. 

\end{abstract}


%
\IEEEpeerreviewmaketitle

\section{Introduction}

Sequence data modeling became an important task in the field of machine learning, finding applications in a wide range of areas. These include speech recognition \cite{graves2013speech}, video processing \cite{zhao2018hsa}, biosignal anlaysis \cite{singh2018classification}, and natural language processing \cite{bahdanau2014neural}. Multiple methods directed at solving sequence data analysis tasks were proposed to date. Notable approaches include those based on Recurrent Neural Networks, such as Gated  Recurrent Units \cite{bahdanau2014neural} or Long Short Term Memory models \cite{hochreiter1997long} that aim to explicitly model the sequential nature of the data with variable length and capture its temporal information. In addition, methods based on Transformers have been proposed as well, modelling the data representations as token sequences with self-attention between tokens being the main driving force of the model \cite{vaswani}. Besides, methods that were originally developed for other types of data, such as Convolutional Neural Networks or Neural Bag of Features \cite{nbof, tnbof}, were shown beneficial in sequential data analysis tasks. 

Concurrent with the development of these methods, approaches directed towards improving robustness of baseline models have been emerging, with the attention modules \cite{vaswani} being one of the most notable ones. The goal of attention module is generally defined as highlighting relevant information in the model while suppressing less relevant one. This idea has been applied to a wide range of base models, and explicit definitions of different attention variants vary between specific models and data types. In CNNs, attention is generally calculated in a form  of a learned mask of weights that is applied  element-wise to the intermediate feature  representation to facilitate learning of stronger features, where mask can be applied both in channel or spatial dimensions \cite{hu2018squeeze, woo2018cbam, ignore}.
Another relevant incarnation of an attention model is that of multi-head self-attention that serves as a building block in Transformer models. In this formulation, relevance of features is quantified by their relations in the learnt latent space. 

Bag of Features \cite{lazebnik2006beyond} model has been widely used for feature extraction from image data, later emerging to other data types as well, including sequential data \cite{jiang2007towards, riley2008text}. The learning process of BoF consists of two stages, with the first stage being dictionary learning, during which a codebook of representative features (codewords) is learnt. During the second stage of BoF, the learnt codebook is used to quantize the low-level feature representation of data into a histogram. To facilitate more powerful feature extraction, Discriminant Bag of Features approaches were proposed \cite{dmbows,dbows}, while Neural Bag of Features (NBoF) was proposed as a neural network generalization of BoF \cite{nbof}. NBoF can be used as an independent feature extractor or as a submodule of a bigger architecture, and can be optimized end-to-end in either case. Besides, an attention module for Neural Bag of Features has been recently proposed to address some of its limitations and increase the robustness of the model \cite{2da, 9130881}. Specifically, 2DA proposed three attention types: input attention, with the aim of addressing the noise present in input data; codeword attention, with the aim of highlighting most relevant codewords in a codebook; and temporal attention, with the aim of highlighting most relevant temporal dimensions in the representation. 

In this paper, we propose to reformulate the idea of 2D-Attention in sequence data and evaluate it in 
Neural Bag of Features model. Our  contributions are summarized as follows: 
\begin{itemize}
    \item We revisit the definition of 2D-Attention, and propose self-attention based alternatives capable of more powerful quantification of feature relevance. We propose self-attention based formulations of both temporal and codeword attention.
    \item We develop codeword-temporal self-attention to facilitate learning of representation relevance in joint codeword-temporal latent space, rather than treating codeword and temporal attentions separately.
    \item We evaluate the developed methods on sequence data analysis tasks, including acoustic scene classification and cardiac disease recognition from ECG and PCG signals, and aciheve competitive performance.
\end{itemize}

The remainder of the paper is organized as follows. Section \ref{S:RelatedWork} provides an overview of the related work, Section \ref{S:ProposedMethods} describes the proposed formulations of 2DA self-attentions, Section \ref{S:Experiments} provides experimental results evaluating their performance in a variety of time-series analysis problems against related approaches, and Section \ref{S:Conclusion} concludes the paper.

\section{Related Work}\label{S:RelatedWork}

Neural Bag of Features (NBoF) \cite{passalis2017neural} is a neural extension of the Bag of Features algorithm that can be utilized both as an independent learning method, as well as incorporated into larger models to facilitate more powerful feature extraction. NBoF consists of two steps, namely, dictionary learning and feature quantization. Specifically, NBoF model receives as input a variable-size representation and quantizes it into a fixed-size histogram representation. Quantization is performed using a learned dictionary that can be optimized jointly with the full model architecture in an end-to-end manner. Further, aggregation step is performed, where the extracted histogram representations, known as codewords, are aggregated by averaging. To date, several feature quantization approaches have been proposed, including those based on Radial Basis Function (RBF) \cite{passalis2017neural} and hyperbolic kernel \cite{tnbof}. Here we revisit the original definition based on RBF kernel.

Formally, NBoF with an RBF kernel is defined as follows. Given a sequence of $N$ feature representations $\mathbf{X} = [\mathbf{x}_1, \dots, \mathbf{x}_N] \in \mathbb{R}^{D \times N}$, the quantization layer produces a sequence of quantized features $\boldsymbol{\Phi} = [\boldsymbol{\phi}_1, \dots, \boldsymbol{\phi}_N] \in \mathbb{R}^{K\times N}$, where $\boldsymbol{\phi}_n = [\phi_{n, 1}, \dots, \phi_{n, K}]^T \in \mathbb{R}^K$ is the quantized representation corresponding to feature $\mathbb{x}_n$. The output of $k^{th}$ RBF neuron for feature $\mathbf{x}_n$ is given as follows:
\begin{equation}\label{eq1}
\phi_{n, k} = \frac{\textrm{exp}\big(-\|(\mathbf{x}_n - \mathbf{v_k})\odot \mathbf{w}_k \|_2\big)}
{\sum_{m=1}^{K}\textrm{exp}\big(-\|(\mathbf{x}_n - \mathbf{v}_m) \odot \mathbf{w}_m \|_2 \big)},
\end{equation}
where $\mathbf{v}_k$ is the $k^{th}$ codeword, $K$ is the total number of codewords, and  $\mathbf{w}_k\in \mathbb{R}^D$ is a learnable parameter controlling the shape of the Gaussian kernel. 

Following the quantization step, the quantized features are aggregated by averaging:
\begin{equation}\label{eq2}
\mathbf{y} = \frac{1}{N} \sum_{n=1}^{N} \boldsymbol{\phi}_n.
\end{equation}

Although providing reasonable feature extraction capabilities in a variety of problems, NBoF has a number of limitations. One of such limitations is that each learned codeword is considered to be equally important in the learned representation, and hence contributes equally to the prediction, although it is reasonable to assume that certain codewords have learnt more powerful features. With respect to sequence data modeling, another limitation is that during the aggregation step, quantized features are combined by simple averaging, disregarding the relative importance of each timestamp. Nevertheless, temporal information can be of great importance in a variety of sequence learning tasks, such as speech command recognition, or dynamic activity recognition, where order of learnt feature representation can be a defining factor for the prediction.

To address these limitations, an attention mechanism for sequence data has been proposed with NBoF as a baseline in mind \cite{2da}. Specifically, the method is referred to as 2D-Attention (2DA) and defines three attention types: input attention, codeword attention, and temporal attention, that aim to emphasize the most relevant input data features, quantized features, and temporal timestamps, respectively.  

Formally, 2DA is defined as follows. Given a feature representation $\bm{\Phi}$, 2DA learns an attention matrix $\mathbf{A}$:
\begin{equation}\label{eq:tau-att1}
 A = softmax(\bm{\Phi} \mathbf{W}),
\end{equation}
where $softmax(\cdot)$ function is applied row-wise to encourage competition between columns of $\bm{\Phi}$, and $\mathbf{W}$ is a learnable weight matrix with diagonal elements fixed at $\frac{1}{N}$. The learnt attention matrix is subsequently applied as:
\begin{equation}\label{eq4}
\tilde{\bm{\Phi}} = \mathcal{F}_{2DA}(\bm{\Phi}) = \alpha (\bm{\Phi} \odot \mathbf{A}) + (1-\alpha) \bm{\Phi},
\end{equation}
where $\alpha$ is a learnt parameter controlling the strength of attention matrix and $\tilde{\bm{\Phi}}$ is the attended representation. 

The first attention type introduced in 2DA is the codeword attention, the aim of which is to highlight most relevant codewords obtained at quantization step of the NBoF model while suppressing the non-relevant ones. This is desirable under the assumption that the output of each quantization neuron contributes differently to the final prediction. Formally, given the output of the quantization step $\bm{\Phi} \in \mathbb{R}^{K\times N}$, an attention mask $\mathbf{A} \in \mathbb{R}^{K \times N}$ of attention weights is applied to the features $\bm{\Phi}$ in order to highlight or suppress its rows, i.e., codewords, by applying the 2DA to $\bm{\Phi}^T$:
\begin{equation}
 \tilde{\bm{\Phi}}_{CA} = \mathcal{F}_{2DA}(\bm{\Phi}^T).
\end{equation}

Similarly, 2DA can be applied directly on the input of NBoF rather than its quantized output in order to improve the robustness of the model towards noise. Since it is desired to highlight individual series in the input data, the process is similar to that of codeword attention, and 2DA is applied to $\mathbf{X}^T$:
\begin{equation}
 \tilde{\mathbf{X}}_{IA} = \mathcal{F}_{2DA}(\mathbf{X}^T).
\end{equation}
This type of attention is referred to as input attention.

In turn, temporal attention aims to highlight relevant timestamps in the sequence during the aggregation step of the NBoF model to address the limitation of the representations being simply averaged during the aggregation step. Formally, it is achieved by applying 2DA on columns of $\bm{\Phi}$: 
\begin{equation}
 \tilde{\bm{\Phi}}_{TA} = \mathcal{F}_{2DA}(\bm{\Phi}).
\end{equation}

\section{Proposed methods}\label{S:ProposedMethods}

Although the 2DA attention addresses certain limitations of the NBoF model in terms of highlighting most relevant attributes in the quantized feature representation, further improvement can be achieved by reformulating the attention learning methodology. 

One limitation of previously proposed 2DA attention mechanism is that attention is applied separately to either codebook or temporal dimensions. Even if both attention masks are learnt and applied simultaneously, such approach does not take into account potential relationships of learned codewords with the temporal representations in the training phase as the masks are learned independently. At the same time, they are not necessarily independent in real-world problems, as certain codewords can have different importance at different timestamps. We therefore hypothesize that learning of joint codeword-temporal attention map can be beneficial for learning better feature representations and therefore assist in classification task.

\subsection{Codeword-temporal self-attention}
Formally, we define the codeword-temporal attention as follows, building on top of the well studied self-attention module. Considering a NBoF-learned feature representation $\bm{\Phi} \in \mathbb{R}^{K\times N}$, where $K$ denotes the number of codewords and $N$ denotes the temporal length, we obtain the attention matrix by quantifying the relations between codeword and temporal features in a joint learnt space. Formally, we define two learnable projection matrices $\mathbf{W}^n_q \in d \times N$, $\mathbf{W}^n_k \in d \times K$ and project the representation $\bm{\Phi}$ temporally and codeword-wise into a joint $d$-dimensional space. 
\begin{eqnarray}
 \mathbf{q}_n = & \bm{\Phi} {\mathbf{W}^n_q}^T,\:\:\: \mathbf{q}_n \in K \times d,\nonumber \\
 \mathbf{k}_n = & \bm{\Phi}^T {\mathbf{W}^n_k}^T,\:\:\: \mathbf{k}_n \in N \times  d.
\end{eqnarray}

Further, to quantify the relations of learnt features in the joint space we calculate the scaled dot-product similarity between representations learned from  temporal dimension and the ones learned from the codebook and apply an activation function $\sigma$, to scale the values. Since at this time we do not aim to promote competition within codewords or timestamps, but rather learn a joint two-dimensional attention matrix, we choose the sigmoid activation function to scale the values to desirable range. An alternative can be using softmax over flattened 2D representation, but we empirically observed no benefit in following this approach. Further, the learnt attention matrix is applied element-wise to  the input feature representation. Following the widely-used definition of multi-head self-attention \cite{vaswani}, $n$ attention matrices can be calculated independently, with the outputs of all heads subsequently concatenated. The attention matrix $\mathbf{A}_n$ corresponding to the head $n$, feature representation $\tilde{\bm{\Phi}}_n$, and the combined feature representation $\tilde{\bm{\Phi}}$ are therefore given as:
 \begin{equation}\label{eq:tau-att}
 \mathbf{A}_n = \sigma(\frac{\mathbf{q}_n\mathbf{k}_n^T}{\sqrt{d}}) \in K \times N,
  \end{equation} 
  \begin{equation}
        \tilde{\bm{\Phi}}_n  = \alpha_n \bm{\Phi} + (1-\alpha_n) \mathbf{A}_n \odot \bm{\Phi}, \\
  \end{equation}
  \begin{equation}
  \tilde{\bm{\Phi}}  = [\tilde{\bm{\Phi}}_1, ..., \tilde{\bm{\Phi}}_n].
  \end{equation}  

\subsection{Codeword self-attention}
 
  A similar idea can be further developed into enhancing the independent codebook and temporal attentions in 2DA. In the standard definition, the projection matrix $\mathbf{W}$ outlined in Eq.~\ref{eq:tau-att1} is fully learnt from scratch, with a role of highlighting relevant codewords or temporal features in $\bm{\Phi}$. Although by design the aim of $\mathbf{W}$ is to converge to the values that reflect the relevance of the corresponding codewords/timestamps, being optimized from scratch, nothing ensures or guides $\mathbf{W}$ towards reflecting these relations. To account for this, we propose to explicitly derive the attention matrix by means of calculating dot-product similarity of codewords in the latent space. That is, considering the codeword attention, we define two learnable projection matrices $\mathbf{W}^n_q \in d \times N$ and $\mathbf{W}^n_k \in d \times N$ from which latent representations of $
 \bm{\Phi}$ are learnt as:
\begin{eqnarray}
 \mathbf{q}_n = & \bm{\Phi} {\mathbf{W}^n_q}^T \in K \times d, \nonumber \\
 \mathbf{k}_n = & \bm{\Phi} {\mathbf{W}^n_k}^T \in K \times  d.
\end{eqnarray}
Following this, we can calculate the codeword attention as a $K \times K$ matrix following Eq.~\ref{eq:tau-att}, where we utilize softmax as $\sigma$ to promote competition  between codewords. 

Note that unlike 2DA, following this approach the learnable parameters are responsible for merely learning a latent space, where relevance of the codewords is explicitly calculated by means of dot product similarity, rather than directly learning the relevance of each codeword as in 2DA. 
The learnt attention matrix is subsequently multiplied with a feature representation $\bm{\Phi}$ to highlight the most relevant codewords and multi-head approach can be followed here as well:
  \begin{equation}
  \tilde{\bm{\Phi}}_n  = \alpha_n \bm{\Phi} + (1-\alpha_n) \mathbf{A}_n \bm{\Phi} \\
  \end{equation}
  \begin{equation}
  \tilde{\bm{\Phi}}  = [\tilde{\bm{\Phi}}_1, ..., \tilde{\bm{\Phi}}_n].
  \end{equation}  

\subsection{Temporal self-attention}
Following the same principle, temporal self-attention can be defined by quantifying temporal relevance of the representation by calculating this in a latent space. To achieve this, temporal self-attention can be calculated by simply operating on the transpose of the feature representation $\bm{\Phi}$, leading to $N \times N$ attention matrix encoding relative importance of each temporal dimension. Specifically, the queries, keys, and combined multi-head representation can be achieved as follows:
 \begin{eqnarray}
  \mathbf{q}_n = & \mathbf{\Phi}^T {\mathbf{W}^n_q}^T \in N \times d \nonumber \\
  \mathbf{k}_n = & \mathbf{\Phi}^T {\mathbf{W}^n_k}^T \in N \times  d
 \end{eqnarray}
  \begin{equation}
  \tilde{\bm{\Phi}}_n  = \alpha_n \bm{\Phi}^T + (1-\alpha_n) A_n \bm{\Phi}^T \\
  \end{equation}
 \begin{equation}
  \tilde{\mathbf{\Phi}} = [(\tilde{\mathbf{\Phi}}_1)^T, ..., (\tilde{\mathbf{\Phi}}_n)^T]
 \end{equation}

\section{Experimental evaluation}\label{S:Experiments}

In this section we report the experimental evaluation of the proposed self-attention mechanisms and compare them with standard 2-DA attention. All the experiments are conducted with the logistic formulation of Neural Bag of Features \cite{nbof} that uses hyperbolic kernel as a quantization layer and we use 256 codewords. We perform experiments on two tasks, namely, biosignal analysis and audio analysis. We denote by 2DA-CA and 2DA-TA the conventional 2DA attention in its codebook and temporal formulations, respectively, and by 2DA-CTSA$_d$, 2DA-TSA$_d$, and 2DA-CSA$_d$ - the proposed variants of codebook-temporal self-attention, temporal self-attention, and codebook self-attention with the dimensionality of the latent space denoted by $d$. Note that $d$ is a hyperparameter which can be tuned, but we instead report the results across multiple values. Unless otherwise specified, single-head models are used.

\subsection{Audio analysis}

The first type of sequence data that we consider is audio. Specifically, we evaluate the NBoF models with the proposed attention approaches on the task of acoustic scene classification defined by TUT-UAS2018 dataset \cite{acoustic}. The dataset poses a task of classification of surrounding environments by their sounds, where 10 classes of urban environments are defined: airport, shopping mall, metro station, street pedestrian, public square, street traffic, tram, bus, metro, park. We extract mel-spectrogram feature representations with 128 frequency bands that are used as an input to a set of convolutional layers as defined in \cite{2da} to facilitate feature extraction, followed by NBoF module. The models are trained for 90 epochs with Adam optimizer and we use the batch size of 256.  
We utilize accuracy as the performance metric and report the accuracy of validation set on 90th epoch averaged across three runs. 

The results of the proposed attention models and competing standard 2DA models are reported in Table I. Here and throughout the paper, we highlight the best result in bold and underline the results that outperform the baseline 2DA attention models. Specifically, TSA, i.e., temporal self-attention is compared with TA, i.e., standard temporal attention, CSA is compared with CA, and CTSA is considered to outperform standard 2DA if it outperforms both CA and TA, i.e., both standard codeword and temporal attention models. 

As can be seen in Table I, the best result is achieved by the proposed temporal self-attention model that outperforms both 2DA baselines. All of the proposed temporal self-attention models outperform the temporal 2DA, and similar result is achieved by codeword self-attention that mostly outperforms codeword 2DA. All of the variants of the proposed codeword-temporal attentions outperform the baseline 2DA. 

\begin{table}[]\caption{Accuracies on TUT-UAS2018 dataset}
\small
\centering
\begin{tabular}{|l|c|}
\hline
\textbf{Attention models}   & \textbf{TUT-UAS}            \\ \hline
2DA-CA       & 56.15 + 0.21        \\
2DA-TA       & 56.09 + 0.51        \\ \hline
2DA-CTSA$_{512}$ &  {\ul 56.20 + 1.11}      \\ 
2DA-CTSA$_{256}$ & {\ul 57.53 + 1.28}      \\
2DA-CTSA$_{128}$ & {\ul 56.84 + 1.07}     \\
2DA-CTSA$_{64}$ & {\ul 56.56 + 0.59}      \\\hline
2DA-TSA$_{512}$   & {\ul 56.81 + 0.63}       \\ 
2DA-TSA$_{256}$   & {\ul \textbf{57.55 + 1.40}} \\
2DA-TSA$_{128}$   & {\ul 57.11 + 0.86}      \\
2DA-TSA$_{64}$    & {\ul 56.91 + 0.91}      \\\hline
2DA-CSA$_{512}$   & {\ul 57.18 + 0.53}     \\
2DA-CSA$_{256}$   &  55.62 + 1.11      \\
2DA-CSA$_{128}$   & {\ul56.94 + 0.62}       \\
2DA-CSA$_{64}$    & {\ul 56.74 + 1.44}      \\
\hline
\end{tabular}
\end{table}

\subsection{Biosignal analysis}
The second type of sequence data considered by our approach is biosignal data. Timely diagnosis of potential heart abnormalities, such as atrial fibrillation or other cardiovascular diseases is an important problem in the modern world, with a multitude of solutions proposed to address it. In our experiments addressed towards this task, we consider two of the widely-adopted biosignals, namely, Electrocardiogram (ECG) and Phonocardiogram (PCG). The first dataset that we consider is the Atrial Fibrillation dataset (AF) that poses the task of atrial fibrillation recognition from ECG signals which are provided as the
development data (training set) in the Physionet/Computing in Cardiology Challenge 2017 \cite{af}. Specifically, the task is formulated as a classification problem with 4 classes:  normal sinus rhythm, atrial fibrillation, alternative rhythm, and noise. Each ECG signal lasts between 9 to 60 seconds, which we clip or zero-pad to achieve the length of 30 seconds. Further, prior to applying the NBoF module, we add several preprocessing convolutional layers to the model to facilitate feature extraction. Specifically, we utilize the same architecture as proposed in \cite{2da}. We perform 5-fold cross-validation and report the average F1 score across the folds. The rest of training hyperparameters are as described in audio classification task.

The second dataset considered for the task of biosignal analysis is the PCG dataset of Phonocardiograms that come
from the training set provided in the Physionet/Computing in Cardiology Challenge 2016 \cite{pcg}. Two different tasks are posed in this dataset: abnormal phonocardiogram detection, and phonocardiogram quality evaluation, where both tasks are binary classification problems. Due to varying lengths of signals in the datset, we extract 5 second segments for classification similarly to \cite{2da}. For feature preprocessing, we extract mel-spectrogram with 24 bands and a window of 25 ms, which are subsequently fed to several preprocessing convolutional layers similarly to \cite{2da} and then to NBoF model. Other training hyperparameters are similar to those of AF dataset, except 3-fold cross-validation is used due to the smaller daatset size. 

The results of biosignal analysis tasks are shown in Table II. As can be seen, in all three cases the best result is achieved by one of the proposed variants. In PCG dataset, codeword-temporal variant in high dimensions outperforms both codeword and temporal 2DA, and codeword self-attention significantly outperforms the codeword 2-DA. At the same time, in temporal representations the proposed approach outperform the 2DA approach in quality evaluation task on PCG dataset. Similar results are observed in AF dataset, where proposed self-attention approaches outperform codeword and temporal 2DA.

\begin{table}[]\caption{F1 scores on biosignal datasets}
\small
\centering
\begin{tabular}{|l|ccc|}
\hline
\textbf{Attention models} & \textbf{PCG}        & \textbf{PCG-2}            & \textbf{AF}    \\ \hline
2DA-CA            & 86.93   + 0.35       & 73.44   + 1.23   & 77.33   + 2.44    \\
2DA-TA            & 87.45 + 0.74         & 73.39 + 1.16     & 76.71 + 2.06      \\ \hline
2DA-CTSA$_{512}$  & {\ul 87.75 + 0.78}   & {\ul 73.75 + 1.81}          & {\ul 77.56 + 1.75}     \\
2DA-CTSA$_{256}$  & {\ul 87.46 + 1.30}   & {\ul 73.50 + 0.77}           & {\ul 77.55 + 2.42}      \\
2DA-CTSA$_{128}$  & {\ul 87.74 + 0.65}         & {\ul 73.62 + 1.80}     &  76.96 + 1.24   \\
2DA-CTSA$_{64}$   &  87.07 + 1.02  & 73.38 + 1.36     & {\ul 77.87 + 1.71}  \\ \hline
2DA-TSA$_{512}$   & {\ul 88.06 + 0.61}         & {\ul 73.46 + 1.45}   & {\ul 76.86 + 2.34}    \\
2DA-TSA$_{256}$   & 87.26 + 0.52         & {\ul 74.14 + 1.77}   & {\ul 76.87 + 1.86}   \\
2DA-TSA$_{128}$   & 87.08 + 1.00         & {\ul \textbf{74.47 + 1.03}} & {\ul 77.27 + 2.13}   \\
2DA-TSA$_{64}$    & {\ul 87.77 + 0.61}         & 73.31 + 1.58   & {\ul 76.99 + 1.74}      \\ \hline
2DA-CSA$_{512}$   & {\ul 88.36 + 0.22} & 73.35 + 1.15  & 77.28 + 1.60   \\
2DA-CSA$_{256}$   & {\ul \textbf{88.38 + 0.55}}   & {\ul 73.95 + 0.90}     & {\ul 77.70 + 1.90} \\
2DA-CSA$_{128}$   & {\ul 87.19 + 0.98}   & 73.02 + 2.14  & {\ul \textbf{78.70 + 1.50}}      \\
2DA-CSA$_{64}$    & {\ul 87.71 + 0.44}   & 72.79 + 0.67     & {\ul 77.96 + 1.88} \\ \hline  
\end{tabular}
\end{table}

We further perform evaluation of the proposed methods with respect to different parameters. Specifically, we evaluate utilization of different number of heads in the models, as well as different Neural Bag of Features formulations.

\subsection{Self-attention with multiple heads}

Using multiple heads in self-attention modules has been shown beneficial in a variety of tasks, as learning multiple latent spaces in parallel allows the model to jointly attend to information from different representation subspaces at different positions \cite{vaswani}. On the other hand, using multiple heads yields additional model parameters. Here, we evaluate the proposed self-attention modules with variants consisting of 2 and 4 heads. In these variants, we use dropout of 0.2 on the attention matrix of codeword and temporal formulations as defined in \cite{vaswani}.

Table III shows the results on TUT-UAS dataset using 2 and 4 heads in multi-head self-attention. As can be seen, the proposed approaches mostly outperform the standard 2DA. Compared to single-head variant, the multihead model with 4 heads perform the best, leading to performance gain of tup to 2.5\%. In terms of biosignal datasets shown in Table IV, it can be seen that the overall results are rather similar between the head numbers in terms of which variants perform well in which datasets. In addition, utilization of multiple heads bring an improvement similarly to the acoustic scene dataset. 

\begin{table}[]\caption{Accuracies on TUT-UAS2018 dataset  with 2 and 4 heads}
\small
\centering
\begin{tabular}{|l|cc|}
\hline
\textbf{Attention models} & \textbf{TUT-UAS, h=2}      &  \textbf{TUT-UAS, h=4} \\ \hline
2DA-CA         & 56.15 + 0.21     & 56.15 + 0.21      \\
2DA-TA         & 56.09 + 0.51     & 56.09 + 0.51      \\ \hline
2DA-CTSA$_{512}$   & {\ul 57.23 + 1.00} & {\ul 57.04 + 0.80}    \\
2DA-CTSA$_{256}$   & {\ul 56.15 + 1.17}  & {\ul \textbf{58.52 + 0.70}}   \\
2DA-CTSA$_{128}$   & {\ul \textbf{57.48 + 0.64}}   & {\ul 58.02 + 0.45}   \\
2DA-CTSA$_{64}$    & 54.91 + 1.22 & {\ul 58.07 + 1.92}   \\
2DA-CTSA$_{32}$    & {\ul 57.21 + 0.29}  & {\ul 56.31 + 0.74}   \\ \hline
2DA-TSA$_{512}$     & {\ul 56.61 + 0.91}  &  55.82 + 1.18   \\
2DA-TSA$_{256}$     & 55.80 + 0.98  & {\ul 56.07 + 0.48}   \\
2DA-TSA$_{128}$     & 55.84 + 0.51  & {\ul 57.62 + 1.71}   \\
2DA-TSA$_{64}$      & {\ul 57.83 + 0.16}  & {\ul 56.71 + 0.81}    \\
2DA-TSA$_{32}$      & {\ul 56.31 + 1.10}  & {\ul 57.83 + 0.23} \\ \hline
2DA-CSA$_{512}$     & {\ul 57.40 + 0.23}  &  {\ul 57.13 + 1.20}    \\
2DA-CSA$_{256}$     & {\ul 56.37 + 1.24}  & 55.45 + 0.71   \\
2DA-CSA$_{128}$     & 55.62 + 1.18  & {\ul 56.91 + 1.31}   \\
2DA-CSA$_{64}$      & {\ul 56.99 + 1.21}  & {\ul 56.54 + 1.45}   \\
2DA-CSA$_{32}$      & {\ul 56.04 + 1.06}     & {\ul 56.26 + 1.53}   \\ \hline          
\end{tabular}
\end{table}

\begin{table}[]\caption{F1 scores on biosignal datasets with 2 heads}
\small
\centering
\begin{tabular}{|l|ccc|}
\hline
  \textbf{Models}    & \textbf{PCG-1}  & \textbf{PCG-2}  & \textbf{AF}   \\
\hline
2DA-CA       & 86.93 + 0.35    & 73.44 + 1.23    & 77.33 + 2.44   \\
2DA-TA       & 87.45 + 0.74      & 73.39 + 1.16      & 76.71 + 2.06     \\ \hline
2DA-CTSA$_{512}$ & {\ul 87.80 + 0.73}     &   {\ul 74.57 + 1.14}   &    76.43 + 2.78   \\
2DA-CTSA$_{256}$ &{\ul 87.84 + 0.12}     &  73.14 + 0.70     & {\ul 76.60 + 1.70}     \\
2DA-CTSA$_{128}$ & 87.24 + 0.74     & {\ul 73.77 + 1.02}     & {\ul 77.06 + 1.39}      \\
2DA-CTSA$_{64}$  &{\ul 88.04 + 0.52}          &  {\ul 73.73 + 1.16}   & 76.98 + 1.92 \\
2DA-CTSA$_{32}$  & {\ul 87.63 + 0.83}           & {\ul 73.69 + 0.98}    & {\ul 77.79 + 2.00}     \\ \hline
2DA-TSA$_{512}$   & 86.98 + 0.76          & {\ul 73.66 + 0.78}      &  {\ul 76.77 + 2.26}     \\
2DA-TSA$_{256}$   & {\ul 87.61 + 0.70}      &  73.32 + 1.15      &  76.31 + 1.59    \\
2DA-TSA$_{128}$   & {\ul 87.69 + 1.11}      & 72.64 + 2.19 & {\ul 77.23 + 1.48}    \\
2DA-TSA$_{64}$    & 87.09 + 0.60     & {\ul 73.55 + 0.80}     & {\ul 77.21 + 1.95}    \\
2DA-TSA$_{32}$    & 87.03 + 0.44       & {\ul 74.38 + 1.81}       & {\ul 77.41 + 2.18}    \\ \hline
2DA-CSA$_{512}$   & {\ul 87.47 + 0.78}      &   72.97 + 0.72    & {\ul \textbf{77.88 + 1.43}}   \\
2DA-CSA$_{256}$   & {\ul \textbf{88.31 + 0.60}}     & {\ul \textbf{74.94 + 1.77}}      & {\ul 77.70 + 1.69}    \\
2DA-CSA$_{128}$   & {\ul 87.33 + 0.68}      & {\ul 74.46 + 0.62}     & {\ul 77.47 + 0.96}    \\
2DA-CSA$_{64}$    & {\ul 87.49 + 0.84}      & 73.41 + 1.13      & 76.91 + 1.11    \\
2DA-CSA$_{32}$    & {\ul 87.72 + 0.57}     & 73.08 + 0.60     & 76.69 + 1.22 \\ \hline
\end{tabular}
\end{table}

\begin{table}[]
\caption{F1 scores on biosignal datasets with 4 heads}
\small
\centering
\begin{tabular}{|l|ccc|}
\hline
\textbf{Attention models} & \textbf{PCG-1}  & \textbf{PCG-2}  & \textbf{AF }  \\
\hline
2DA-CA    & 86.93   + 0.35    & 73.44   + 1.23    & 77.33   + 2.44    \\
2DA-TA    & 87.45 + 0.74      & 73.39 + 1.16      & 76.71 + 2.06      \\ \hline
2DA-CTSA$_{512}$        &  {\ul 87.88 + 0.56}     &  {\ul 74.34 + 0.85}    &  77.09 + 1.24   \\
2DA-CTSA$_{256}$        & {\ul 88.04 + 0.53}      &   {\ul 73.37 + 1.94} & {\ul \textbf{78.26 + 1.69}}    \\
2DA-CTSA$_{128}$        & 86.91 + 0.29     & 72.60 + 1.18 & {\ul 77.51 + 1.75}    \\
2DA-CTSA$_{64}$         & 86.65 + 0.57     & {\ul 73.62 + 1.58}    &  {\ul 77.87 + 2.29}    \\
2DA-CTSA$_{32}$         & {\ul 87.74 + 0.67}     & 73.12 + 0.13      & {\ul 77.61 + 1.66}      \\ \hline
2DA-TSA$_{512}$          &  86.94 + 0.60      &   72.92 + 1.76    &   76.69 + 1.54   \\
2DA-TSA$_{256}$          & {\ul 87.45 + 0.71}      & {\ul 73.59 + 0.89}     & {\ul 77.13 + 1.84}     \\
2DA-TSA$_{128}$          & 87.20 + 0.23      & 73.37 + 0.80     & {\ul 77.03 + 2.18}    \\
2DA-TSA$_{64}$           & 87.29 + 0.39      & {\ul 74.07 + 1.11}     & {\ul76.94 + 2.47}     \\
2DA-TSA$_{32}$           & 87.03 + 0.90      & 73.69 + 0.73      & {\ul 77.56 + 1.80}     \\ \hline
2DA-CSA$_{512}$          & {\ul 88.27 + 0.63} & 73.29 + 1.51       & {\ul 77.77 + 1.72}  \\
2DA-CSA$_{256}$         & {\ul \textbf{88.59 + 0.81}}     & {\ul 74.16 + 1.58}      & {\ul 77.59 + 1.64}     \\
2DA-CSA$_{128}$          & {\ul 87.67 + 0.41}     & 72.82 + 1.82      & {\ul 77.30 + 1.60}     \\
2DA-CSA$_{64}$           & {\ul 87.37 + 0.56}     & {\ul 73.60 + 1.40}     & 77.26 + 1.67     \\
2DA-CSA$_{32}$           & {\ul 87.12 + 0.66}     & {\ul \textbf{74.35 + 0.97}}     & 76.73 + 1.77    \\ \hline
\end{tabular}
\end{table}

\subsection{Using other NBoF formulations}

All experiments were performed with the logistic NBoF formulation \cite{nbof}. However, our proposed self-attention module can be equally utilized with other formulations as well. Here, we evaluate our approaches and competing 2DA approaches using the temporal variant of NBoF that defines two codebooks, long-term and short-term \cite{tnbof}. We refer to this method as TNBoF. Here we utilize the acoustic scene classification dataset and evaluate the TNBoF baseline with our approaches with both single and multi-head variants, as shown in Tables VI and VII. We can see that using this model, codeword self-attention mostly outperforms the basedline codeword 2DA, and the other variants mostly outperform the baseline. At the same time, it can be seen that overall the best performing variant is the single-head one, hence utilization of additional heads degrades the performance rather than improves it in this case. 

\begin{table}[]
\caption{Accuracy scores on TUT-UAS2018 datast with TNBoF model with 1 head}
\small
\centering
\begin{tabular}{|l|c|}
\hline
 \textbf{Attention models}  & \textbf{TUT-UAS, TNBoF }\\ \hline
2DA-CA       & 56.79 + 0.60         \\
2DA-TA       & 55.89 + 0.34        \\ \hline
2DA-CTSA$_{64}$  &  {\ul  57.04 + 0.84}      \\
2DA-CTSA$_{128}$ & {\ul  56.51 + 0.46} \\
2DA-CTSA$_{256}$ & {\ul  57.35 + 1.05}     \\
2DA-CTSA$_{512}$ & {\ul  \textbf{58.19 + 0.62}}     \\ \hline
2DA-TSA$_{64}$    & {\ul 56.94 + 0.63}      \\
2DA-TSA$_{128}$   &  {\ul 56.46 + 0.63}      \\
2DA-TSA$_{256}$   &  {\ul 56.32 + 0.26}      \\
2DA-TSA$_{512}$   &  {\ul 56.09 + 0.62}     \\ \hline
2DA-CSA$_{64}$    & 55.70 + 0.20        \\
2DA-CSA$_{128}$   & 56.41 + 0.44        \\
2DA-CSA$_{256}$   & {\ul 56.83 + 0.57}        \\
2DA-CSA$_{512}$   &  56.46 + 0.29      \\ \hline
\end{tabular}
\end{table}

\begin{table}[]\caption{Accuracy scores on TUT-UAS2018 dataset with TNBoF model with 2 and 4 heads, respectively}
\small
\centering
\begin{tabular}{|l|c|c|}
\hline
\textbf{Attention models} & \multicolumn{2}{c|}{\textbf{TUT-UAS, TNBoF}}            \\ \hline
& \multicolumn{1}{c|}{h=2} & h=4 \\ \hline
2DA-CA         &      56.79   + 0.60 & 56.79   + 0.60    \\
2DA-TA         &       55.89 + 0.34 & 55.89 + 0.34    \\ \hline
2DA-CTSA$_{512}$   & 56.26 + 0.92   &    56.36 + 0.82    \\
2DA-CTSA$_{256}$   & {\ul 57.08 + 0.86}      & 56.57 + 0.65   \\
2DA-CTSA$_{128}$   & {\ul \textbf{57.77 + 0.91}} & {\ul 57.31 + 1.01}       \\
2DA-CTSA$_{64}$   & {\ul 56.51 + 1.89}      & 56.42 + 0.22      \\
2DA-CTSA$_{32}$  & {\ul 57.25 + 0.91}   & {\ul 57.38 + 0.79}   \\ \hline
2DA-TSA$_{512}$     & {\ul 56.51 + 1.03}   & {\ul 56.79 + 0.59 }  \\
2DA-TSA$_{256}$   & {\ul 57.45 + 0.97}   & {\ul 56.98 + 1.79}    \\
2DA-TSA$_{128}$    & {\ul 56.61 + 1.16}   & {\ul 56.78 + 0.58}   \\
2DA-TSA$_{64}$     & 55.95 + 0.99      & {\ul 56.24 + 1.24}   \\
2DA-TSA$_{32}$      & {\ul 56.17 + 0.31}  & {\ul \textbf{58.40 + 0.70}} \\ \hline
2DA-CSA$_{512}$    &   55.89 + 0.52     &   56.15 + 1.47   \\
2DA-CSA$_{256}$   & {\ul 57.13 + 1.18}      & 56.29 + 1.97      \\
2DA-CSA$_{128}$    & {\ul 57.04 + 1.11}      & 55.45 + 0.41   \\
2DA-CSA$_{64}$     & 55.77 + 1.58      & 55.95 + 0.91      \\
2DA-CSA$_{32}$      & 55.60 + 0.68     & {\ul 57.09 + 0.71}      \\ \hline          
\end{tabular}
\end{table}

\section{Conclusion}\label{S:Conclusion}

In this paper, we revisited the standard formulation of a 2DA attention mechanism and proposed several ways of enhancing it. The proposed ways are based on self-attention and allow to quantify codeword and/or temporal relevances through latent spaces rather than learning them directly. We evaluated the proposed approaches together with the Neural Bag-of-Features model on a few sequence learning tasks. The experimental evaluation has shown the benefits of the proposed approaches. Since the proposed attention models are generic methods aimed towards multivariate sequence data, further work into its applications with other architectures and tasks remains as a future research direction.  

\section{Acknowledgement} 
This project has received funding from the European
Union’s Horizon 2020 research and innovation programme
under grant agreement No 871449 (OpenDR). This publication
reflects the authors views only. The European Commission
is not responsible for any use that may be made of the
information it contains.






\bibliographystyle{IEEEtran}
\bibliography{references}
%



\end{document}